\documentclass[lettersize,journal]{IEEEtran}
\usepackage{amsmath,amsfonts}
\usepackage{array}
\usepackage[caption=false,font=normalsize,labelfont=sf,textfont=sf]{subfig}
\usepackage{textcomp}
\usepackage{stfloats}
\usepackage{url}
\usepackage{verbatim}
\usepackage{graphicx}
\usepackage{cite}
\usepackage{enumerate}
\usepackage[backref]{hyperref}
\usepackage{multirow}
\usepackage{multicol}
\usepackage[table,xcdraw]{xcolor}
\usepackage{booktabs}
\usepackage{makecell}
\usepackage[flushleft]{threeparttable}
\usepackage[normalem]{ulem}
\useunder{\uline}{\ul}{}
 \usepackage{xcolor}
 \usepackage{multirow}
\usepackage[table,xcdraw]{xcolor}
\usepackage[flushleft]{threeparttable}
 \usepackage[ruled,linesnumbered]{algorithm2e}
\usepackage{listings}
\usepackage{diagbox}
\usepackage{pifont}
\newcommand{\cmark}{\ding{52}\xspace}
\newcommand{\xmark}{\ding{56}\xspace}

\begin{document}

\title{URA-Net: Uncertainty-Integrated Anomaly Perception and Restoration Attention Network \\ for Unsupervised Anomaly Detection}

\author{
Wei Luo, Peng Xing, Yunkang Cao, Haiming Yao, \\ Weiming Shen,~\IEEEmembership{Fellow,~IEEE,} and Zechao Li$^{*}$,~\IEEEmembership{Senior Member,~IEEE}

\thanks{Wei Luo and Haiming Yao are with the State Key Laboratory of Precision Measurement Technology and Instruments, Department of Precision Instrument, Tsinghua University, Beijing, China (e-mail:luow23@mails.tsinghua.edu.cn; yhm22@mails.tsinghua.edu.cn).}
\thanks{Yunkang Cao and Weiming shen are with the State Key Laboratory of Intelligent Manufacturing Equipment and Technology, Huazhong University of Science and Technology, Wuhan 430074, China (e-mail:
cyk\_hust@hust.edu.cn; wshen@ieee.org).}
\thanks{Peng Xing and Zechao Li are with the School of Computer Science and Engineering, Nanjing University of Science and Technology, Nanjing 210094, China (e-mail: xingp\_ng@njust.edu.cn; zechao.li@njust.edu.cn). \textit{(Corresponding author: Zechao Li.)}}
\thanks{This work was supported by National Natural Science Foundation of China (Grant No. 62425603) and Basic Research Program of Jiangsu Province (Grant No. BK20240011).}
}

\markboth{}%
{Shell \MakeLowercase{\textit{et al.}}: A Sample Article Using IEEEtran.cls for IEEE Journals}


\maketitle
\begin{abstract}
Unsupervised anomaly detection plays a pivotal role in industrial defect inspection and medical image analysis, with most methods relying on the reconstruction framework. However, these methods may suffer from over-generalization, enabling them to reconstruct anomalies well, which leads to poor detection performance. To address this issue, instead of focusing solely on normality reconstruction, we propose an innovative \uline{U}ncertainty-Integrated Anomaly Perception and \uline{R}estoration \uline{A}ttention \uline{Net}work (URA-Net), which explicitly restores abnormal patterns to their corresponding normality. First, unlike traditional image reconstruction methods, we utilize a pre-trained convolutional neural network to extract multi-level semantic features as the reconstruction target. To assist the URA-Net learning to restore anomalies, we introduce a novel feature-level artificial anomaly synthesis module to generate anomalous samples for training. Subsequently, a novel uncertainty-integrated anomaly perception module based on Bayesian neural networks is introduced to learn the distributions of anomalous and normal features. This facilitates the estimation of anomalous regions and ambiguous boundaries, laying the foundation for subsequent anomaly restoration. Then, we propose a novel restoration attention mechanism that leverages global normal semantic information to restore detected anomalous regions, thereby obtaining defect-free restored features. Finally, we employ residual maps between input features and restored features for anomaly detection and localization. The comprehensive experimental results on two industrial datasets, MVTec AD and BTAD, along with a medical image dataset, OCT-2017, unequivocally demonstrate the effectiveness and superiority of the proposed method.
\end{abstract}


\begin{figure}
    \centering
    \includegraphics{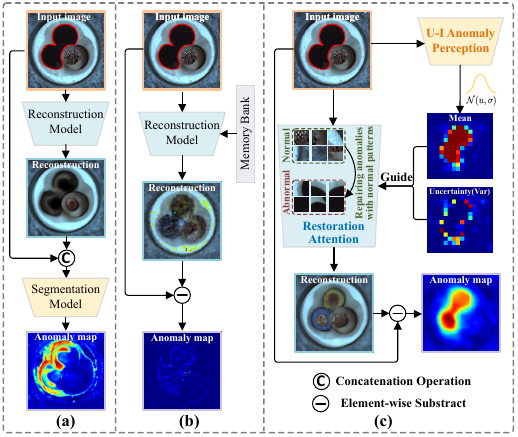}
    \caption{\textbf{Comparison of different unsupervised anomaly detection methods.} (a) DRAEM \cite{draem}. (b) MemAE \cite{MemAE}. (c) The proposed method (URA-Net). URA-Net employs the U-I (Uncertainty-Integrated) Anomaly Perception module to roughly estimate anomalous regions (\textbf{Mean}) and ambiguous boundaries (\textbf{Uncertainty}). Subsequently, The Restoration Attention module utilizes global normal semantic information to restore the detected anomalies, ultimately resulting in defect-free restored images. It is noteworthy that our proposed method relies on feature reconstruction. The reconstructed images are generated by training a decoder, which is exclusively employed for visualization.} 
    \label{fig:intro}
\end{figure}
\section{Introduction}
\IEEEPARstart{I}{mage} anomaly detection is a critical task in the field of computer vision, which aims to determine whether a given sample deviates from the pre-defined normality.  It has garnered increasing attention due to its applications across various domains, including industrial quality control \cite{MVTEC, GLCF, PBAS, tao2023vitalnet, tao2022deep}, medical analysis \cite{medical, fernandes2020automatictnnls}, and video surveillance \cite{video, learnable, influence}. In real-world scenarios, obtaining a substantial number of anomalous samples and annotating them with pixel-level labels is both time-consuming and labor-intensive. This constraint limits the applicability of supervised learning methods \cite{PGANet}. Consequently, we prioritize unsupervised anomaly detection methods, which use solely normal images for training.\\
\indent Currently, a majority of unsupervised anomaly detection methods adhere to the reconstruction \cite{AE, normalitylearning, sa-memory, pmemory, tao2022unsupervised, cao2025varad} framework. This framework is built on the assumption that an established model trained with solely normal samples can only reconstruct normal patterns well but fail on abnormal ones. In this way, the reconstruction error for anomalous patterns will be substantially larger than that for normal patterns, facilitating the distinction between normal and anomalous samples. However, due to over-generalization \cite{caoyunkangtii, cao2024bias} of well-trained neural networks, anomalous patterns may also be reconstructed well. To tackle this issue, some methods \cite{draem, li2021cutpaste, RIAD, NDP-Net, wang2024produce, NIGS} transform the reconstruction task into a restoration task, in which they introduce artificial anomalies to normal samples and enforce the model to restore the normal appearances. For instance, DRAEM \cite{draem} utilizes natural images to synthesize artificial anomalous ones, enabling the reconstruction model to learn to restore these artificial anomalies. Subsequently, a segmentation model is employed for anomaly localization. However, these methods encounter a significant limitation: the reconstruction model lacks an explicit anomaly restoration mechanism. It only relies on minimizing the difference between the input normal image and the reconstructed artificial anomalous one to address anomalies. This approach fails to provide adequate contextual information, often resulting in reconstructed images exhibiting unknown patterns, thereby leading to poor detection performance, as illustrated in Fig. \ref{fig:intro}(a).\\
\indent Therefore, some explicit anomaly restoration mechanisms \cite{MemAE, TrustMAE, Intra, DAAD} have been proposed. For example, MemAE \cite{MemAE} uses an external memory bank to store typical normal prototype features and replaces anomalous features with pre-stored normal features to achieve anomaly restoration. However, these methods have two limitations: i) Using a memory bank to store normal features results in additional memory usage and computational overhead. ii) Not only are abnormal features processed, but also normal features are forced to be replaced by pre-stored prototype features in the memory bank. While this aids in suppressing the reconstruction of abnormal regions, it simultaneously leads to a degradation in the quality of reconstruction in normal regions, thus affecting the detection performance, as depicted in Fig. \ref{fig:intro}(b).\\
\indent To this end, we introduce a novel Uncertainty-Integrated Anomaly Perception and Restoration Attention Network (URA-Net) for unsupervised anomaly detection, as presented in Fig. \ref{fig:intro}(c). The proposed URA-Net more closely aligns with human intuition in restoring anomalous regions, \textit{i.e.}, it first roughly estimates the normal and anomalous regions, then utilizes the normal semantic information from the global context to restore the anomalous regions. This method ensures that the restored anomalous regions closely align with the real distribution while maintaining the original structure of the normal regions, resulting in superior restoration and detection outcomes. Specifically, URA-Net follows the paradigm of feature reconstruction \cite{DFR}, which utilizes a pre-trained convolutional neural network (CNN) to extract multi-level semantic features as the reconstruction targets. To facilitate URA-Net in learning to restore anomalous regions, we propose a Feature-level Artificial Anomaly Synthesis Module (FASM), which aims to generate diverse abnormal samples that differ from the normal pattern to participate in training. Furthermore, an Uncertainty-Integrated Anomaly Perception Module (UIAPM) is introduced, which aims to roughly estimate normal and abnormal regions, laying the groundwork for subsequent anomaly restoration. Within UIAPM, we employ discriminative learning to enhance the model's capacity in discerning normal/abnormal patterns. Moreover, we integrate Bayesian neural networks (BNN) into UIAPM, transitioning it from a point estimation model to a distribution estimation model. We believe this offers two key advantages: i) The distribution estimation model can furnish uncertainty estimates for detection outcomes, thereby assisting the model in identifying ambiguous boundaries. ii) Artificial anomalies generated by FASM might induce overfitting in point estimation models. Specifically, while the model adeptly detects anomalies similar to artificial ones, it encounters difficulties with genuine industrial anomalies that significantly diverge from artificial ones. A distribution estimation model can better mitigate such overfitting tendencies. Following this, we present a Restoration Attention Module (RAM), which utilizes global normal features under the guidance of UIAPM to restore anomalous regions while maintaining the original structure of normal regions without additional memory usage and computational overhead. Therefore, the proposed URA-Net notably enhances anomaly restoration effectiveness and achieves superior anomaly detection capabilities.\\
\indent We analyze the detection performance of URA-Net through extensive experiments conducted on two industrial datasets, MVTec AD \cite{MVTEC} and BTAD \cite{BTAD}, as well as a medical image dataset, OCT-2017 \cite{oct2017}. Our proposed URA-Net demonstrates superior detection performance compared to previous state-of-the-art methods. The main contributions of this study can be summarized as follows:
\begin{itemize}
    \item We propose a novel Restoration Attention Module (RAM) that utilizes global normal semantic information to restore anomalous regions, achieving enhanced anomaly restoration without additional computational overhead. This approach ensures that the restored anomalous regions better conform to the real distribution while preserving the original structure of normal regions.
    \item To facilitate anomaly restoration in RAM, we introduce a novel Uncertainty-Integrated Anomaly Perception Module (UIAPM), which aims to roughly estimate normal and abnormal regions.
    \item To facilitate the model in learning to restore anomalous regions, we propose a Feature-level Artificial Anomaly Synthesis Module (FASM), which aims to generate diverse abnormal samples for training.
\end{itemize}

\section{Related Work}

\begin{figure*}
    \centering
    \includegraphics[width=180mm]{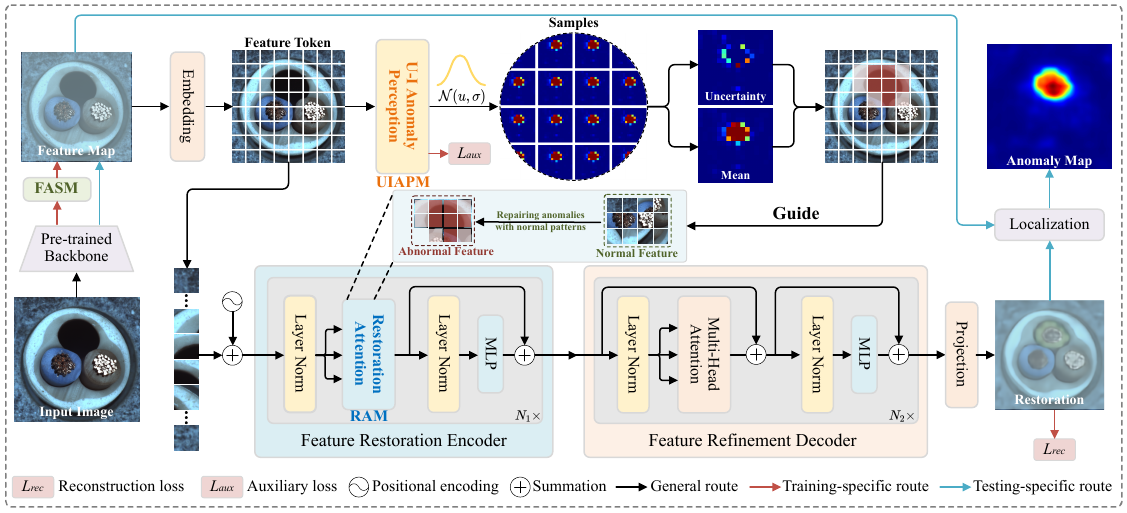}
    \caption{\textbf{Overall architecture of our URA-Net.} It primarily consists of three modules: feature-level artificial anomaly synthesis module (FASM), uncertainty-integrated anomaly perception module (UIAPM), and restoration attention module (RAM). First, a pre-trained backbone is employed to transform input images into multi-level features. FASM is utilized to generate artificial anomalies at the feature level for training. Subsequently, UIAPM roughly estimates anomalous regions (\textbf{Mean}) and ambiguous boundaries (\textbf{Uncertainty}). Then RAM leverages global normal semantic information to restore detected anomalous regions and ambiguous boundaries, yielding defect-free restored features. Finally, the residuals between input features and restored ones are utilized for anomaly detection and localization.}
    \label{fig:URA-Net}
\end{figure*}
\subsection{Unsupervised Anomaly Detection}
This study classifies existing unsupervised anomaly detection methods into two primary categories: embedding-based and reconstruction-based methods.
\subsubsection{Embedding-based Method}
The embedding-based methods employ pre-trained models to extract representations of normal images, subsequently compressing these representations into a specific embedding space. Within this space, normal features tend to aggregate together, while abnormal features are notably distant from the normal clusters. During testing, the distance between test features and normal cluster serves as the criterion for anomaly detection. For example, Deep SVDD \cite{DeepSVDD} constructs a hypersphere to delineate the boundary between normal and abnormal data. Within the hypersphere lie normal samples, while abnormal samples exist outside of it, thereby achieving image anomaly detection. To achieve anomaly localization, Patch SVDD \cite{PatchSVDD} extends Deep SVDD to the patch level. However, the detection efficiency of Patch SVDD is low because it requires extracting a large number of patches. To improve the computation speed, GCPF \cite{GCPF} models normal samples using a multivariate Gaussian distribution and employs the Mahalanobis distance as an anomaly score. MBPFM \cite{MBPFM} employs two different pre-trained networks to extract features, which are then mutually mapped. Precise anomaly localization is achieved by calculating the mapping errors between them. However, these methods suffer from overfitting due to the high generalization capability of neural networks. To mitigate this overfitting, CDO \cite{caoyunkangtii} collaboratively optimizes the distributions of both normal and abnormal features. To further enhance performance, PatchCore \cite{PacthCore} utilizes a greedy coreset subsampling algorithm to construct a memory bank containing typical normal features. During testing, the distance between test features and pre-stored normal features serves as anomaly scores. 
\subsubsection{Reconstruction-based Method}
The reconstruction-based methods are built on the assumption that models trained on normal samples can only reconstruct normal patterns and cannot reconstruct abnormal patterns. During testing, the residual image between the input image and its reconstruction is utilized for anomaly detection. Auto Encoder (AE) \cite{AE} is a classic reconstruction model. However, owing to the formidable generalization capacity of neural networks, anomalies can still be perfectly reconstructed during testing. To alleviate this issue, MemAE \cite{MemAE} utilizes a memory bank within the latent feature space to store normal features. To further enhance the performance of anomaly detection, TrustMAE \cite{TrustMAE} combines MemAE with perceptual distance \cite{perceptual_distance}. Additionally, 
Additionally, RIAD \cite{RIAD}, DRAEM \cite{draem}, and NDP-Net \cite{NDP-Net} methods generate artificial anomalies to train the model to effectively suppress the reconstruction of anomalies. Due to the absence of semantic information in individual pixel values, DFR \cite{DFR} employs pre-trained networks to extract multi-level semantic features as reconstruction targets, thus achieving enhanced detection performance. To strike a fine balance between detection accuracy and inference speed, some methods \cite{PNPT, chen2022utrad, GLCF} develop Transformer-based frameworks for anomaly detection. To better suppress the reconstruction of anomalous features, AMI-Net \cite{AMI-Net} introduces an adaptive mask generator to effectively conceal anomalous regions. Moreover, FOD \cite{fod} further enhances the detection performance of feature reconstruction methods through intra- and inter-correlation learning. However, these methods do not explicitly utilize global semantic information to restore anomalous regions, leading to poor anomaly restoration quality. To this end, we propose URA-Net, which first roughly estimates normal and anomalous regions, and then uses global normal semantic information to restore the anomalous regions. This approach enhances both the quality of anomaly restoration and the performance of anomaly detection.
\subsection{Bayesian Neural Networks}
Bayesian neural networks (BNN) can enhance the ability to capture the uncertainty inherent in model parameters through the integration of probability distributions over weights \cite{blundell2015weight} and features \cite{feature2019}. There have been numerous works incorporating BNN into vision perception tasks. For instance, UGTR \cite{uncertaintyCOD} integrates BNN with Vision Transformer (ViT) \cite{Vit} to enhance the accuracy of camouflage object detection. UC-Net \cite{zhang2021uncertaintyRBGD} employs uncertainty for RGB-D saliency detection. Upformer \cite{huang2022pixeluncer} leverages a memory-enhanced encoder and an uncertainty-aware decoder to enhance the robustness of supervised surface defect segmentation. Inspired by these works, we propose an uncertainty-integrated anomaly perception module (UIAPM) as a probabilistic model to capture uncertainty for unsupervised anomaly detection.
\section{Proposed Method}
\subsection{Problem Formulation}
In unsupervised anomaly detection and localization, a model is optimized using a training set comprising only normal samples and then evaluated on a test set containing both normal and anomalous samples. The training dataset with $N$ normal samples is denoted as $\mathcal{D}_{train}=\{I_{n}^{(i)}\}_{i=1}^{N}$, where $I_{n}^{(i)}$ denotes the $i^{th}$ normal image. The test dataset with $M$ samples is denoted as $\mathcal{D}_{test}=\{(I_{t}^{(i)},y_{t}^{(i)},m_{t}^{(i)})\}_{i=1}^{M}$, where $I_{t}^{(i)}$ denotes the $i^{th}$ test image with its image label $y_{t}^{(i)}\in\{0, 1\}$ and pixel-wise label $m_{t}^{(i)}$. Here, 0 indicates normal, and 1 indicates anomalous. The objective is to utilize $\mathcal{D}_{train}$ to establish a mapping $f:I_t^{(i)} \to (y_{t}^{(i)},m_{t}^{(i)})$ for the detection and localization of anomalies in $\mathcal{D}_{test}$.

\subsection{Model Overview}
Fig. \ref{fig:URA-Net} presents the framework of URA-Net, which comprises three main modules: the feature-level artificial anomaly synthesis module (FASM), the uncertainty-integrated anomaly perception module (UIAPM), and the restoration attention module (RAM). Initially, following the feature reconstruction method proposed by DFR \cite{DFR}, we utilize a pre-trained backbone to transform input images into multi-scale semantic features (\textit{Sec. \ref{Sec_backbone}}). To aid the anomaly restoration process, we introduce FASM (\textit{Sec. \ref{Sec_FASM}}). During training, FASM generates artificial anomalies at the feature level. During inference, the multi-level features are directly forwarded into subsequent networks without FASM. Subsequently, we employ the UIAPM (\textit{Sec. \ref{Sec_UIPAM}}) to roughly estimate anomalous features (\textbf{Mean} in Fig. \ref{fig:URA-Net}) and ambiguous boundaries (\textbf{Uncertainty} in Fig. \ref{fig:URA-Net}), which lay the foundation for subsequent anomaly restoration. Then, under the guidance of UIAPM, the RAM (\textit{Sec. \ref{Sec_RAM}}) leverages the global normal semantic information to restore detected anomalous regions. A feature refinement decoder is then employed to refine the details of features, resulting in defect-free restored features. Ultimately, the discrepancies between the input features and the restored features are employed for anomaly detection and localization.

\subsection{Multi-level Feature Extraction}
\label{Sec_backbone}
Feature-level reconstruction \cite{DFR} has been proven to be a better strategy in comparison to image-level reconstruction.   
Therefore, we utilize a backbone $\phi$ pretrained on ImageNet \cite{ImageNet} to extract multi-scale features $\{\phi_{1}(I), \phi_{2}(I), \cdots ,\phi_{l}(I)\}$ from input images $I$. Considering the multi-scale nature of real anomalies, we fuse features from multiple levels. Specifically, we use interpolation to resize features from different levels to a uniform size and subsequently concatenate them along the channel dimension to obtain multi-level fused features $F(I)$, which are then utilized as the reconstruction targets.
\begin{equation}
     F(I) = \mho \{\Gamma (\phi_{1}(I)), \Gamma (\phi_{2}(I)), \cdots ,\Gamma (\phi_{l}(I))\}
\end{equation}
where $F(I)\in \mathbb{R}^{H_F\times W_F \times C_F}$, $\Gamma$ denotes the scaling operation, and $\mho$ represents the concatenation operation along the channel dimension.
\begin{figure}
    \centering
    \includegraphics[width=87mm]{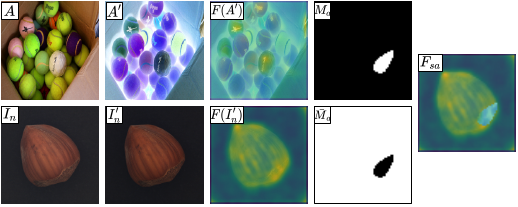}
    \caption{\textbf{Synthesis process for feature-level artificial anomalies.}}
    \label{fig:FASM}
\end{figure}
\subsection{Feature-level Artificial Anomaly Synthesis Module}
\label{Sec_FASM}
We present the FASM to generate synthetic anomalies, thereby aiding the learning process for anomaly restoration.
In contrast to DRAEM \cite{draem}, which generates artificial anomalies at the image level, FASM operates at the feature level. This approach can effectively reduce the impact of noise and enhance the model's robustness to feature perturbations. \\
\indent As depicted in Fig. \ref{fig:FASM}, initially, we obtain a normal image $I_n$ from the training set, and an anomaly source image $A$ is sampled from the ImageNet \cite{ImageNet} dataset, possessing a distribution entirely distinct from that of the input image. Then, we independently perform data augmentation operations on $I_n$ and $A$ to derive the enhanced images, denoted as $I_n^{\prime}$ and $A^{\prime}$. 
It is noteworthy that the data augmentation operations applied to $A$ include $\{$\textit{posterization}, \textit{sharpness}, \textit{solarization}, \textit{equalization}, \textit{brightness variation}, \textit{color variation}, \textit{contrast variation}$\}$. Conversely, the data augmentation operations applied to $I_n$ consist solely of $\{$\textit{brightness variation}, \textit{contrast variation}$\}$. Subsequently, we utilize the pre-trained backbone to extract multi-level fused features $F(I_n^{\prime})$ and $F(A^{\prime})$ from images $I_n^{\prime}$ and $A^{\prime}$, respectively, as described in \textit{Sec. \ref{Sec_backbone}}. A random mask $M_a\in\mathbb{R}^{H_F\times W_F \times 1}$ capable of simulating various anomaly shapes is generated using a Perlin noise generator \cite{perlin1985image}. Finally, the synthesized  artificial anomaly features $F_{sa}$ are generated by combining $F(I_n^{\prime})$, $F(A^{\prime})$, and $M_a$:
\begin{equation}
    F_{sa} = \bar{M}_{a}\odot F(I_{n}^{\prime })+M_{a}\odot F(A^{\prime} )
\end{equation}
where $F_{sa}\in \mathbb{R}^{H_F\times W_F \times C_F}$, $\bar{M}_{a}$ represents the inverse of $M_a$, and $\odot$ denotes the element-wise multiplication operation. In the following sections, we utilize $F_n$, $F_a$, and $F_{sa}$ to respectively represent normal features, real-world anomaly features, and synthesized artificial anomaly features for clarity. Here, $F_a$ is equivalent to $F(I_a)$, where $I_a$ denotes a real anomalous image from the test set.\\
\indent Through FASM, the URA-Net can adapt to various anomalies at the feature level, enhancing its ability to comprehend and address diverse real-world anomalies.
\subsection{Uncertainty-Integrated Anomaly Perception Module}
\label{Sec_UIPAM}
\begin{figure}
    \centering
    \includegraphics[width=85mm]{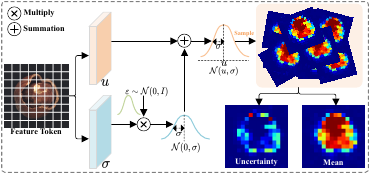}
    \caption{\textbf{Illustration of uncertainty-integrated anomaly perception module (UIAPM).} UIAPM works as a probabilistic model to roughly estimate anomalous regions and ambiguous boundaries.}
    \label{fig:uiapm}
\end{figure}

To lay the foundation for subsequent anomaly restoration in RAM, we propose UIAPM, which roughly estimates anomalous regions and ambiguous boundaries. UIAPM integrates two key techniques: discriminative learning and BNN.\\
\indent To ease the computation complexity of UIAPM, we convert the feature maps into feature token sequences. As depicted in Fig. \ref{fig:URA-Net}, an embedding layer $f_{embed}$ with a convolutional kernel of size $K$ is employed to convert the 2D multi-scale fused features $F_{sa}$ into a 1D feature token sequence $E_{sa}=\{T_{sa}^1, T_{sa}^2, \cdots, T_{sa}^L|T_{sa}^i\in \mathbb{R}^D\}$.
\begin{equation}
    E_{sa} = f_{embed}(F_{sa}; \theta_{embed})
\end{equation}
where $E_{sa}\in \mathbb{R}^{L\times D}$, with $L=\frac{H_F}{K}\times\frac{W_F}{K}$ denoting the number of feature tokens, $D$ signifies the dimensionality of feature channels, $f_{embed}$ and $\theta_{embed}$ respectively represent the function and parameters of the embedding layer.\\
\indent As depicted in Fig. \ref{fig:uiapm}, instead of merely predicting a fixed scalar, UIAPM integrates BNN to transform it from a point estimation model into a distribution estimation model, thereby obtaining the probability distribution of detection results for each feature token (\textit{e.g.} the token $T_{sa}^i$). Following previous works \cite{kendall2015bayesian, uncertaintyCOD, huang2022pixeluncer}, we define the anomaly score $z_{sa}^i\in \mathbb{R}^1$ generated by UIAPM for feature token $T_{sa}^i$ as a Gaussian distribution:
\begin{equation}
    z_{sa}^i\sim \mathcal{N}(u_i, \sigma_i^2)
\end{equation}
where the mean $u_i$ and standard deviation $\sigma_i$ of the Gaussian distribution are input-dependent generated by two linear layers ($f_u$ and $f_{\sigma}$) within UIAPM.
\begin{equation}
\begin{aligned}
    u_i = f_u(T_{sa}^i;\theta_{u}),\quad \sigma_i = f_{\sigma}(T_{sa}^i;\theta_{\sigma})
\end{aligned}
\end{equation}
where $\theta_u$ and $\theta_\sigma$ respectively represent the model parameters with regard to output $u_i$ and $\sigma_i$.\\
\indent Therefore, the anomaly score is no longer a fixed scalar value, but rather a random value sampled from $\mathcal{N}(u_i, \sigma_i^2)$. However, the random sampling operation is non-differentiable, thereby hindering gradient back-propagation. 
To address this issue, inspired by variational autoencoders \cite{VAE}, we employ the re-parameterization strategy. Specifically, we sample a random value $\varepsilon $ from a standard Gaussian distribution $\mathcal{N}(0,I)$, and then obtain the sampled anomaly score as follows. 
\begin{equation}
\label{Eq:sample}
z_{sa}^{i}=u_{i}+\varepsilon\sigma_{i}
\end{equation}
In this manner, the gradients can be propagated
backward to optimize the model parameters. $z_{sa}^i$ represents the final anomaly score for feature token $T_{sa}^i$. With $L$ tokens, we obtain a sequence of anomaly scores $Z_{sa}^{(m)}\in \mathbb{R}^{L\times1}$ ({$Z_{sa}^{(m)}$ denotes a sample randomly drawn from the learned distribution, where $m$ represents the $m$-th sampling iteration) for the entire feature token sequence $E_{sa}$. We can then apply a max-pooling operation with a patch size of $K$ to $M_a$ and then flatten it to obtain the corresponding ground truth $G_{sa}\in \mathbb{R}^{L\times1}$. Binary cross-entropy (BCE) is used to measure the discrepancy between the prediction $Z_{sa}^{(m)}$ and true label $G_{sa}$.\\
\indent Furthermore, to better distinguish normal and abnormal instances, we adopt a discriminative learning strategy. During training, we utilize paired inputs: normal feature token sequence $E_n$ and synthesized artificial anomaly feature token sequence $E_{sa}$. Hence, {the discriminative loss $\mathcal{L}_{dis}$ is defined as follows:
\begin{equation}
\label{Eq:L_contra}
\mathcal{L}_{dis}=\mathcal{L}_{BCE}(Z_{sa}^{(m)}, G_{sa})+\mathcal{L}_{BCE}(Z_{n}^{(m)}, G_{n})
\end{equation}}
where $Z_{n}^{(m)}$ and $G_n$ represents the prediction and ground truth of $E_{n}$, respectively. {The generation process of $Z_{n}^{(m)}$ and $G_n$ is fully aligned with that of $Z_{sa}^{(m)}$ and $G_{sa}$.} To promote diversity and accelerate the training process, we solely utilize one sample for computing the loss. \\
\indent However, \begin{figure}
    \centering
    \includegraphics[width=82mm]{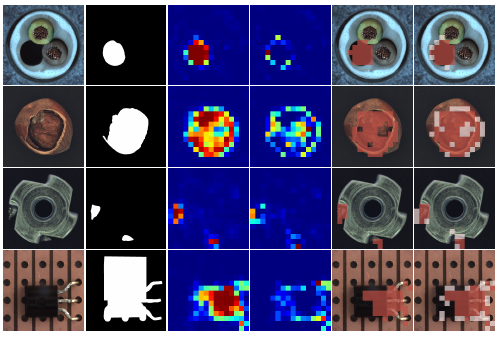}
    \caption{\textbf{The effectiveness of UIAPM in anomaly perception.} From left to right are: anomalous image, ground truth, mean map $U$, uncertainty map $V$, mask $M_U$ generated by $U$, and the final mask $M_{final}$ combined with uncertainty.}
    \label{fig:uncer_vis}
\end{figure}solely optimizing UIAPM using the loss function in Eq. \ref{Eq:L_contra} would lead to the $\sigma_i$ in Eq. \ref{Eq:sample} approaching $0$. Therefore, the stochastic output can be reformulated as $z_{sa}^i=u_i+0$, degrading to the deterministic output. To address this issue, we explicitly constrain $\mathcal{N}(u_i, \sigma_i^2)$ to approach the standard Gaussian distribution $\mathcal{N}(0, I)$. The Kullback-Leibler (KL) divergence is employed to quantify the disparity between $\mathcal{N}(u_i, \sigma_i^2)$ and $\mathcal{N}(0, I)$.
\begin{equation}
\begin{aligned}
     \mathcal{L}_{kl} & = D_{KL}(\mathcal{N}(u, \sigma^2)||\mathcal{N}(0, I))\\
     & =-\frac{1}{2}(1+\log\sigma^2-u^2-\sigma^2)
\end{aligned}
\end{equation}
where $u \in \mathbb{R}^{L\times 1}$ and $\sigma 
 \in \mathbb{R}^{L\times 1}$ represent the mean and standard deviation sequences derived from the UIAPM. The final auxiliary loss function $\mathcal{L}_{aux}$ for optimizing the UIAPM is defined as follows:
\begin{equation}
    \mathcal{L}_{aux} = \mathcal{L}_{dis} + \lambda \mathcal{L}_{kl}
\end{equation}
where $\lambda$ represents a trade-off factor, which is set to 0.001 in our study.\\
\indent During testing, when the real-world anomaly feature token sequence $E_a$ is fed into the UIAPM, we can sample $M$ random anomaly score sequences from the learned probability distribution, represented as $Z_{a}=\{Z_{a}^{(1)}, Z_{a}^{(2)}, \cdots, Z_{a}^{(M)}\}$. More accurate anomaly scores and uncertainty estimates are obtained by calculating the mean and standard deviation of $Z_{a}$.
\begin{equation}
\begin{aligned}
    U = Mean(Z_{a}), \quad V = Var(Z_{a})
\end{aligned}
\end{equation}
where $U\in \mathbb{R}^{L\times 1}$ and $V\in \mathbb{R}^{L\times 1}$ denote the mean sequence and uncertainty sequence. $Mean(\cdot)$ and $Var(\cdot)$ represent the operations for computing the mean and standard deviation, respectively.\\
\indent Subsequently, binary operations are applied separately to $U$ and $V$ to obtain their respective masks, denoted as $M_U\in \mathbb{R}^{L\times 1}$ and $M_V\in \mathbb{R}^{L\times 1}$. We illustrate the binary process by describing the generation of $M_U = \{M_{U}^1, M_{U}^2, \cdots, M_{U}^L|M_{U}^i\in \mathbb{R}^1\}$.
\begin{equation}
    \begin{array}{l} 

 M_{U}^i = \left\{\begin{matrix} 
  1,U^{i}\leq  \lambda_{u}  \\ 
  0,U^{i}> \lambda _{u}
\end{matrix}\right. , \quad \lambda_{u} = \varpi(U)+\gamma\sigma(U) 
\end{array} 
\end{equation}
where $\varpi(\cdot)$ and $\sigma(\cdot)$ denote the calculation of the mean and standard deviation, respectively, and {$\gamma$ represents the scaling factor used to control the threshold $\lambda_{u}$ for binarization. In our study, $\gamma$ is set to 1.0.}\\
\indent As depicted in Fig. \ref{fig:uncer_vis}, the mask $M_U$ (penultimate column in Fig. \ref{fig:uncer_vis}) generated by $U$ fails to detect all abnormal regions. We attribute this to two reasons: i) the presence of ambiguous boundaries between normal and abnormal patterns, and ii) disparities between the synthesized artificial anomaly distribution during training and the real anomaly distribution. To tackle this issue, we merge the $M_V$ and $M_U$, obtaining the final mask $M_{final}$ (last column in Fig. \ref{fig:uncer_vis}) to detect as many abnormal regions as possible.
\begin{figure}
    \centering
    \includegraphics[width=88mm]{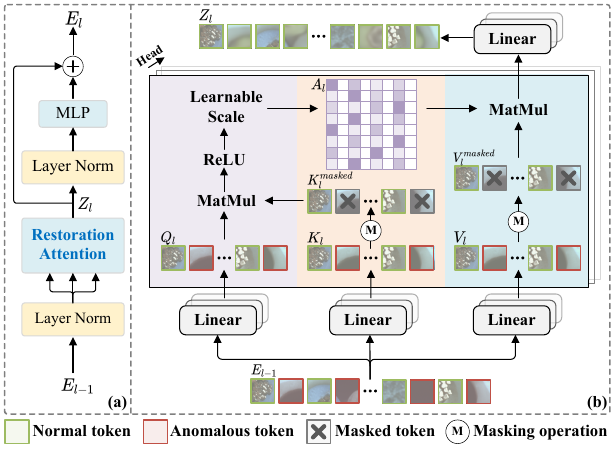}
    \caption{(a) The proposed restoration transformer block consists of three components: layer normalization (LN), \textbf{restoration attention module (RAM)}, and multi-layer perceptron (MLP). (b) The RAM architecture. RAM leverages global normal semantic information to restore anomalous features.}
    \label{fig:RAM}
\end{figure}
\subsection{Restoration Attention Module}
\label{Sec_RAM}
The quality of anomaly restoration significantly impacts anomaly detection performance. Therefore, inspired by the Spa-former \cite{huang2024sparse}, we introduce a novel restoration transformer block that leverages global normal semantic information to restore anomalous regions, thereby improving the quality of anomaly restoration and detection performance.\\
\indent The architecture of the restoration transformer block is depicted in Fig. \ref{fig:RAM}(a). Unlike conventional vision transformer blocks, two significant alterations are made: i) we remove the first residual connection commonly present in vision transformer blocks, as this connection can directly introduce abnormal features into subsequent processes. The reason for retaining the second residual connection is to accelerate model convergence. ii) we introduce a novel restoration attention module (RAM) to replace the original self-attention mechanism. This modification is motivated by the fact that in the original self-attention mechanism, abnormal features tend to correlate with themselves or neighboring abnormal features, leading to their easy and perfect reconstruction.\\
\indent As depicted in Fig. \ref{fig:RAM}(b), we provide a detailed explanation of the mechanism of RAM using the $l$th layer as an example. Given the output token $E_{l-1}\in \mathbb{R}^{L\times D}$ from the previous layer, we initially employ three linear layers to acquire the $Q_{l}$ (query), $K_{l}$ (key), and $V_{l}$ (value), respectively.
\begin{equation}
    [Q_l, K_l, V_l]=E_{l-1}[W_{l}^{Q}, W_{l}^{K}, W_{l}^{V}]
\end{equation}
where $Q_l, K_l, V_l\in\mathbb{R}^{L\times D}$. $W_{l}^{Q}, W_{l}^{K}, W_{l}^{V} \in \mathbb{R}^{D \times D}$ denote the learnable parameters for linear layers.\\
\indent In the attention mechanism, the similarity between 
$Q_{l}$ and $K_{l}$ determines the weight distribution of $V_{l}$, indicating which $V_{l}$ will be assigned higher weights. However, in traditional self-attention mechanisms, $K_{l}$ and $V_{l}$ still contain abnormal information, leading to the perfect reconstruction of abnormal features. Therefore, to mitigate this issue, we need to construct Key-Value pairs without abnormal information. To achieve this, we utilize the mask $M_{final}$ generated by UIAPM (\textit{Sec. \ref{Sec_UIPAM}}) to perform masking operations on $K_{l}$ and $V_{l}$.
\begin{equation}
      [K_l^{masked}, V_l^{masked}] =M_{final}\odot[K_l,V_l]
\end{equation}
where $K_l^{masked}, V_l^{masked} \in \mathbb{R}^{L\times D}$ denote the masked Key and Value. Subsequently, the restored features $Z_l$ are obtained by performing a weighted sum of the masked value $V_l^{masked}$ based on the similarity between $Q_l$ and $K_l^{masked}$.
\begin{equation}
    A_{l} = \beta ReLU(Q_l({K_l^{masked}})^T),\quad Z_l = A_lV_l^{masked}
\end{equation}
where $A_l\in \mathbb{R}^{L\times L}$ denotes the attention map. $(\cdot)^{T}$ denotes the transpose operation. $ReLU(\cdot)$ represents the Rectified Linear Unit (ReLU) activation function. The purpose of using the ReLU activation function instead of the softmax activation function in the traditional self-attention mechanism is to enhance the model's ability to concentrate attention on highly correlated values and to mitigate the impact of weakly correlated values on the attention map. $\beta$ is a learnable scaling factor utilized to adaptively adjust the values in the attention map, ensuring training stability.\\
\indent As depicted in Fig. \ref{fig:RAM}(a), $Z_l$ is subsequently passed into layer normalization (LN) and multi-layer perceptron (MLP) to enhance the feature representation.
\begin{equation}
    E_l = Z_l+MLP(LN(Z_l))
\end{equation}
where $E_l\in \mathbb{R}^{L\times D}$ is the final output of the restoration transformer block.\\
\indent As illustrated in Fig. \ref{fig:URA-Net}, the entire reconstruction network comprises a feature restoration encoder and a feature refinement decoder. The feature restoration encoder consists of $N_1$ restoration transformer blocks, while the feature refinement decoder comprises $N_2$ vanilla transformer blocks. In our study, both $N_1$ and $N_2$ are set to 2.
\subsection{Training Loss and Anomaly Score}
\subsubsection{Training Loss}
Regarding feature reconstruction, we introduce a reconstruction loss function that simultaneously considers local information and global structural information. In the local loss function, we employ both mean squared error (MSE) and cosine similarity to measure the similarity between features, comprehensively considering the direction and magnitude of feature vectors. Specifically, given a normal feature $F_n\in \mathbb{R}^{H_F\times W_F \times C_F}$, a corresponding synthesized artificial anomaly feature $F_{sa}\in \mathbb{R}^{H_F\times W_F \times C_F}$ is generated using FASM (\textit{Sec. \ref{Sec_FASM}}). Subsequently, $F_{sa}$ is fed into URA-Net to obtain the reconstructed feature $\hat{F}_{sa}\in \mathbb{R}^{H_F\times W_F \times C_F}$, ideally restored to $F_{n}$. Therefore, the reconstruction loss function $\mathcal{L}_{rec}$ is defined as follows:
\begin{equation}
    \mathcal{L}_{local}^{mse}=\frac{1}{H_FW_F}\sum_{h=1}^{H_F}\sum_{w=1}^{W_F}{|| F_n(h,w)-\hat{F}_{sa}(h,w) || }_2^2
\end{equation}
\begin{equation}
    \mathcal{L}_{local}^{cos}=\frac{1}{H_FW_F}\sum_{h=1}^{H_F}\sum_{w=1}^{W_F}1-\frac{F_n(h,w)^T\cdot \hat{F}_{sa}(h,w)}{|| F_n(h,w) ||\,|| \hat{F}_{sa}(h,w) ||   }
\end{equation}
\begin{equation}
    \mathcal{L}_{global}=1-\frac{vec(F_n)^T\cdot vec(\hat{F}_{sa} )}{||vec(F_n)||\,||vec(\hat{F}_{sa})|| } 
\end{equation}
\begin{equation}
    \mathcal{L}_{rec} = \underbrace{L_{local}^{mse}+L_{local}^{cos}}_{local}+\underbrace{L_{global}}_{global}
\end{equation}
where ${||\cdot||}_2$, $\cdot$, $||\cdot||$, and $vec(\cdot)$ represent the $L_2$ norm, inner product, modulus length, and flatten operation, respectively.\\
\indent To enhance the anomaly perception capability of the model, as discussed in \textit{Sec. \ref{Sec_UIPAM}}, UIAPM is trained with an auxiliary loss $\mathcal{L}_{aux}$, consisting of $\mathcal{L}_{contra}$ and $\mathcal{L}_{kl}$. Thus, the final joint loss $\mathcal{L}_{final}$ is formulated as follows: 
\begin{equation}
    \mathcal{L}_{final} = \mathcal{L}_{rec} + \mathcal{L}_{aux}
\end{equation}
\subsubsection{Anomaly Score}
During testing, an anomalous feature $F_a\in \mathbb{R}^{H_F\times W_F \times C_F}$ is fed into the trained URA-Net to produce the corresponding reconstructed feature $\hat{F}_a \in \mathbb{R}^{H_F\times W_F \times C_F}$. The disparity between the original anomalous feature and the reconstructed feature is then employed as the anomaly score.
\begin{equation}
    AS_{mse}(h,w)={||F_a(h,w)-\hat{F}_a(h,w)||}_2^2
\end{equation}
\begin{equation}
    AS_{cos}(h,w)=1-\frac{F_a(h,w)^T\cdot \hat{F}_a(h,w)}{|| F_a(h,w) ||\,|| \hat{F}_a(h,w) ||}
\end{equation}
\begin{equation}
AS_{final}=\Theta(AS_{mse}\odot AS_{cos})
\end{equation}
where $\Theta$ represents the operation of scaling to the size of the original image. Following \cite{multilevel-recon}, we employ the standard deviation of $AS_{final}$ as the criterion for image-level anomaly detection.
\section{Experiments}
\subsection{Experimental Settings}
\subsubsection{Datasets}
To validate the superiority and generalizability of the proposed URA-Net, we compare its detection performance against other state-of-the-art methods using two publicly available industrial anomaly detection datasets, MVTec AD \cite{MVTEC} and BTAD \cite{BTAD}, as well as a publicly available medical image anomaly detection dataset, OCT-2017 \cite{oct2017}.\\
\indent \textbf{MVTec AD:} The widely used MVTec AD dataset consists of 15 categories of industrial products, including five texture categories and ten object categories. It contains 3,629 normal images for training and 498 normal images along with 1,982 anomalous images for testing.\\
\indent \textbf{BTAD:} The BTAD dataset includes three types of complex industrial products. For training, it offers 1,799 normal images, and for testing, it provides 451 normal images along with 290 anomalous images. The complex texture backgrounds in this dataset make it particularly challenging.\\
\indent \textbf{OCT-2017:} The OCT-2017 dataset is a medical image dataset consisting of four categories: choroidal neovascularization (CNV), diabetic macular edema (DME), Drusen, and normal. The training set includes 26,315 normal images. The testing set contains 250 normal images and 750 anomalous images.
\subsubsection{Implementation details}
The URA-Net is trained using the AdamW \cite{AdamW} optimizer, with a learning rate of 0.001 and a batch size of eight for 400 epochs. The input images are resized to 256$\times$256 pixels, and the feature map size is set to 64$\times$64. The default backbone is WideResNet50\cite{wideresnet}, where features from the 2nd to 4th layers are resized and then concatenated along the channel dimension, resulting in a 1792-channel feature map. The reconstruction network comprises 2 restoration transformer blocks and 2 vanilla transformer blocks, each with a hidden dimension of 768 and 12 attention heads. The default patch size $K$ is set to 4. All experiments are carried out on a system featuring a 12th Gen Intel(R) Core(TM) i7-12700F CPU at 2.10 GHz, paired with an NVIDIA
GeForce GTX 3060 GPU.
\subsubsection{Evaluation metrics}
We utilize the widely adopted Area Under the Receiver Operating Characteristic Curve (AUROC) metric to evaluate anomaly detection at both the image and pixel levels. Additionally, for medical images, we employ average classification accuracy (ACC) and F1-score as evaluation metrics for anomaly detection. Here, ACC and F1-score are defined as follows: $ACC=\frac{TP+TN}{TP+TN+FP+FN}$ and $\text{F1-score}=\frac{2TP}{2TP+FP+FN}$, where $TP$, $TN$, $FN$, and $FP$ represent true positives, true negatives, false negatives, and false positives, respectively. The threshold for calculating the evaluation metrics is selected based on the optimal F1-score.

\begin{table*}[!h]
\centering
\caption{Anomaly detection and localization results in terms of image/pixel level AUROC on \textbf{MVTec AD} dataset \cite{MVTEC}. The best result is in \textbf{bold}, and the second best is \uline{underlined}. $\dag$ denotes the results obtained through our implementation.} 
\label{table:mvtec}
\vspace{-7pt}
\fontsize{10}{14}\selectfont{
\resizebox{\textwidth}{!}{
\begin{tabular}{cc|ccccc|ccc|ccc|>{\columncolor{pink!30}}c}
\toprule[1.0pt]
\multicolumn{2}{c|}{Taxonomy}          & \multicolumn{5}{c|}{Embedding-Based}           & \multicolumn{3}{c|}{Image Reconstruction} & \multicolumn{4}{c}{Feature Reconstruction}                   \\ \midrule
\multicolumn{2}{c|}{Method~$\rightarrow$}            & PaDiM     & PFM       & PatchCore & CDO$^{\dag}$ & ADPS      & RIAD         & DRAEM        & OCR-GAN    & UTRAD     & RD4AD     & FOD$^{\dag}$       &  \\
\multicolumn{2}{c|}{Venue~$\rightarrow$}             & ICPR'21   & TII'22     & CVPR'22    & TII'23 & TNNLS'24    & PR'21         & ICCV'21       & TIP'23      & NN'22      & CVPR'22    & ICCV'23    & \textbf{URA-Net}                         \\
\multicolumn{2}{c|}{Category~$\downarrow$}          & \cite{PaDiM}          & \cite{MBPFM}          & \cite{PacthCore}           &  \cite{caoyunkangtii} &  \cite{ADPS}        & \cite{RIAD}              & \cite{draem}             & \cite{ocr-gan}           & \cite{chen2022utrad}          & \cite{RD4AD}          &  \cite{fod}         &                          \\ \midrule
\multicolumn{1}{c|}{\multirow{5}{*}{\rotatebox{90}{Texture}}} & Carpet     & \uline{99.8}/99.1 & \textbf{100}/99.2  & 98.7/99.0 & 99.5/99.0 & 97.4/\textbf{99.5} & 84.2/94.2    & 97.0/95.5    & 99.4/-     & 96.3/97.3 & 98.9/98.9 & \textbf{100}/99.2  & \uline{99.8}/\uline{99.3}                 \\
                       \multicolumn{1}{c|}{}  & Grid       & 96.7/97.3 & 98.0/98.8 & 98.2/98.7 & \uline{99.9}/99.2 & \textbf{100}/99.2 &99.6/96.3    & \uline{99.9}/\textbf{99.7}    & 99.6/-     & 98.7/97.6 & \textbf{100}/\uline{99.3}  & \textbf{100}/98.7  & \textbf{100}/99.0                 \\
                       \multicolumn{1}{c|}{}  & Leather    & \textbf{100}/99.2  & \textbf{100}/\uline{99.4}  & \textbf{100}/98.3  & \textbf{100}/99.1  & 
            \textbf{100}/\textbf{99.9}  &\textbf{100}/\uline{99.4}     & \textbf{100}/98.6     & \uline{97.1}/-     & \textbf{100}/98.6  & \textbf{100}/\uline{99.4}  & \textbf{100}/99.3 & \textbf{100}/\uline{99.4}                 \\
                       \multicolumn{1}{c|}{}  & Tile       & 98.1/94.1 & 99.6/96.2 & 98.7/95.6 & 98.9/96.9 & 
                       99.8/\textbf{99.6} &98.7/89.1    & 99.6/\uline{99.2}    & 95.5/-     & \uline{99.9}/95.0 & 99.3/95.6 & \textbf{100}/95.8  & \textbf{100}/96.2                \\
                       \multicolumn{1}{c|}{}  & Wood       & 99.2/94.9 & 99.5/95.6 & 99.2/95.0 & 99.5/95.3 & 97.7/\textbf{99.3} & 93.0/85.8    & 99.1/\uline{96.4}    & 95.7/-     & \textbf{99.7}/93.1 & 99.2/95.3 & 99.3/95.0 & \uline{99.6}/95.5                 \\ \midrule
\multicolumn{1}{c|}{\multirow{10}{*}{\rotatebox{90}{Object}}} & Bottle     & \uline{99.9}/98.3 & \textbf{100}/98.4  & \textbf{100}/98.6  & \textbf{100}/\uline{99.3}  & \textbf{100}/\textbf{99.5}  &\uline{99.9}/98.4    & 99.2/99.1   & 99.6/-     & \textbf{100}/95.9  & \textbf{100}/98.7  & \textbf{100}/98.6  & \textbf{100}/98.8                \\
                        \multicolumn{1}{c|}{} & Cable      & 92.7/96.7 & 98.8/96.7 & \uline{99.5}/\uline{98.4} & 97.8/97.8 & 93.7/94.6 & 81.9/84.2    & 91.8/94.7    & 99.1/-     & 98.6/97.3 & 95.0/97.4 & 98.8/98.3 & \textbf{100}/\textbf{99.0}                \\
                         \multicolumn{1}{c|}{} & Capsule    & 91.3/98.5 & 94.5/98.3 & 98.1/98.8 & 91.9/98.7 & 
                         96.0/98.7 &88.4/92.8    & \textbf{98.5}/94.3    & 96.2/-     & 94.3/97.8 & 96.3/98.7 & 97.1/\uline{99.0} & \uline{98.0}/\textbf{99.1}                 \\
                        \multicolumn{1}{c|}{} & Hazelnut   & 92.0/98.2 & \textbf{100}/99.1  & \textbf{100}/98.7  & 99.4/99.2 &
                        99.6/\uline{99.6} &83.3/96.1    & \textbf{100}/\textbf{99.7}     & 98.5/-     & 99.5/98.4 & \uline{99.9}/98.9 & \textbf{100}/99.0  & \textbf{100}/98.9                 \\
                       \multicolumn{1}{c|}{}  & Metal Nut  & 98.7/97.2 & \textbf{100}/97.2  & \textbf{100}/98.4  & 99.2/\uline{98.5} & 
                       99.7/97.5 &88.5/92.5    & 98.7/\textbf{99.5}    & 99.5/-     & 96.2/95.0 & \textbf{100}/97.3  & \textbf{100}/98.1  & \uline{99.8}/98.1                \\
                        \multicolumn{1}{c|}{} & Pill       & 93.3/95.7 & 96.5/97.2 & 96.6/97.4 & \uline{98.6}/\uline{99.0} & 95.3/\textbf{99.3} &83.8/95.7    & \textbf{98.9}/97.6    & 98.3/-     & 94.2/97.5 & 96.6/98.2 & 96.5/98.7 & 97.2/98.7                \\
                        \multicolumn{1}{c|}{} & Screw      & 85.8/98.5 & 91.8/98.7 & 98.1/99.4 &  90.3/99.3 &89.5/98.7 & 84.5/98.8    & 93.9/97.6    & \textbf{100}/-      & 88.3/97.8 & 97.0/\textbf{99.6} & 96.0/99.1 & \uline{98.3}/\uline{99.5}                \\
                       \multicolumn{1}{c|}{}  & Toothbrush & 96.1/98.8 & 88.6/98.6 & \textbf{100}/98.7  & 86.1/99.0 & 95.3/\textbf{99.1} &\textbf{100}/\uline{98.9}     & \textbf{100}/98.1     & 98.7/-     & 78.9/96.2 & \uline{99.5}/\textbf{99.1} & 95.3/98.6 & 98.9/\uline{98.9}                 \\
                       \multicolumn{1}{c|}{}  & Transistor & 97.4/97.5 & 97.8/87.8 & \textbf{100}/96.3  & 99.4/95.8 & 97.6/92.2 &90.9/87.7    & 93.1/90.9    & 98.3/-     & 96.4/94.9 & 96.7/92.5 & \textbf{100}/\textbf{98.9}  & \textbf{100}/\uline{98.1}                \\
                        \multicolumn{1}{c|}{} & Zipper     & 90.3/98.5 & 97.4/98.2 & \uline{99.4}/\uline{98.5} & 98.9/98.4 & \textbf{100}/\textbf{99.6} & 98.1/97.8    & \textbf{100}/\uline{98.8}     & 99.0/-     & 98.6/97.9 & 98.5/98.2 & 98.2/98.1 & 98.7/98.5                 \\ \midrule
\multicolumn{2}{c|}{Mean}              & 95.5/97.5 & 97.5/97.3 & \uline{99.1}/98.1 & 97.3/\uline{98.3} & 97.4/98.1 & 91.7/94.2    & 98.0/97.3    & 98.3/-     & 96.0/96.7 & 98.5/97.8 & 98.7/\uline{98.3} & \textbf{99.4}/\textbf{98.5}  \\ \bottomrule[1.0pt]  
\end{tabular}}}
\end{table*}

\begin{figure*}
    \centering
    \includegraphics{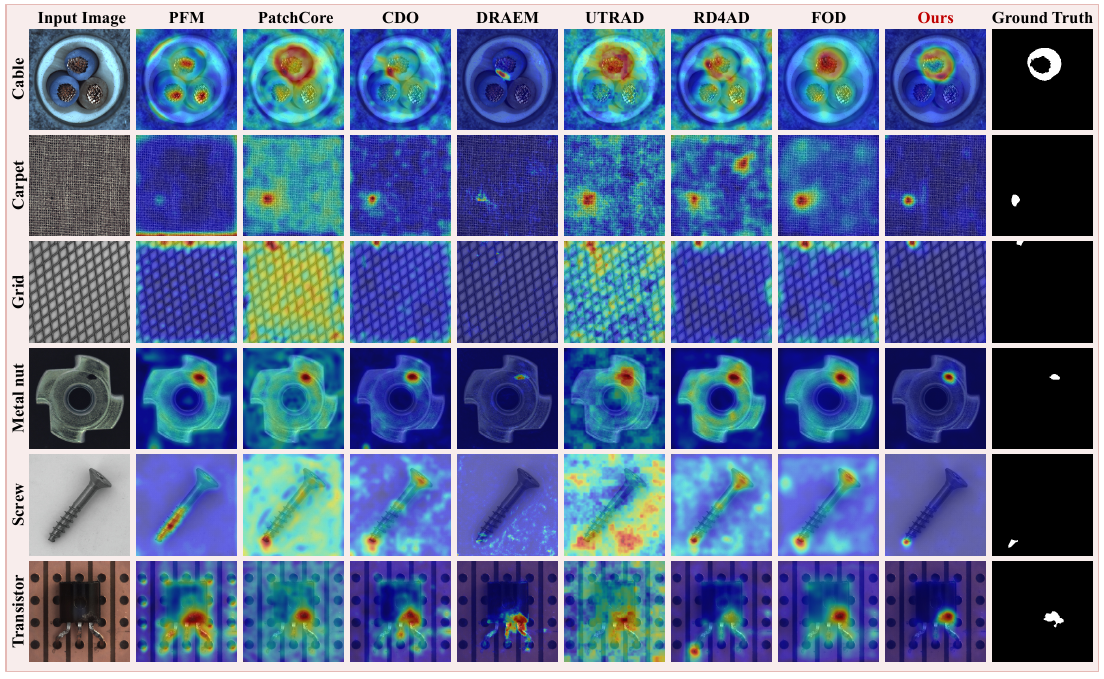}
    \caption{The localization results of the proposed URA-Net and comparative methods on the MVTec AD dataset \cite{MVTEC}. Our proposed model accurately localizes anomaly regions across various categories.}
    \label{fig:mvtec}
\end{figure*}

\begin{table*}[!h]
\centering
\caption{Anomaly detection and localization results in terms of image/pixel level AUROC on \textbf{BTAD} dataset \cite{BTAD}. The best result is in \textbf{bold}, and the second best is \uline{underlined}. $\dag$ denotes the results obtained through our implementation.} 
\label{table:btad}
\vspace{-7pt}
\fontsize{10}{14}\selectfont{
\resizebox{0.95\textwidth}{!}{
\begin{tabular}{c|cccccccccc|>{\columncolor{pink!30}}c}
\toprule[1.0pt]
Method~$\rightarrow$ & PaDiM     & VT-ADL    & FastFlow & MKD       & PFM$^{\dag}$   & PatchCore & DRAEM     & UTRAD$^{\dag}$ & RD4AD$^{\dag}$  & FOD$^{\dag}$    &  \\
Venue~$\rightarrow$     & ICPR'21   & ISIE'2021  & Arxiv'21 & CVPR'21   & TII'22 & CVPR'22    & ICCV'21    & NN'22  & CVPR'22 & ICCV'23 & \textbf{URA-Net}                         \\
Category~$\downarrow$  & \cite{PaDiM}          & \cite{BTAD}          & \cite{fastflow}          & \cite{MKD}          & \cite{MBPFM}      & \cite{PacthCore}           & \cite{draem}           & \cite{chen2022utrad}       & \cite{RD4AD}        & \cite{fod}       &                          \\ \midrule
Product 01 & \uline{99.8}/97.0 & 97.6/\textbf{99.0} & 99.4/97.1      & 93.8/94.9 & 92.9/95.0   & 98.4/\uline{97.3} & 99.5/92.7 & 97.7/93.8   & 98.4/96.3    & 99.6/97.2    & \textbf{99.9}/97.0                      \\
Product 02 & 82.0/96.0 & 71.0/94.0 & 82.4/93.6      & \textbf{88.2}/96.3 & 80.0/\textbf{96.7}   & 81.8/96.1 & 77.4/93.6 & 76.8/95.5   &  86.5/\uline{96.6}   & 86.7/95.7    & \uline{88.1}/96.2                      \\
Product 03 & 99.4/98.8 & 82.6/77.0 & 91.1/98.3      & 98.5/98.3 & 99.6/\uline{99.6}   & \textbf{100}/99.3  & \uline{99.8}/96.4 & 99.7/99.4   & 99.7/99.7   & \textbf{100}/\uline{99.6}    & \textbf{100}/\textbf{99.7}                      \\ \midrule
Mean      & 93.7/97.3 & 83.7/90.0 & 90.1/96.3      & 93.5/96.5 & 90.8/97.1   & 93.4/\textbf{97.6} & 92.2/94.2 & 91.4/96.2   &  94.9/\uline{97.5}   & \uline{95.4}/\uline{97.5}    & \textbf{96.0}/\textbf{97.6}  \\
\bottomrule[1.0pt]  
\end{tabular}}}
\end{table*}

\begin{table*}[!h]
\centering
\caption{Anomaly detection results in terms of image level AUROC on \textbf{OCT-2017} dataset \cite{oct2017}. The best result is in \textbf{bold}, and the second best is \uline{underlined}.} 
\label{table:oct-2017}
\vspace{-7pt}
\fontsize{12}{16}\selectfont{
\resizebox{\textwidth}{!}{
\begin{tabular}{c|ccccccccccc|>{\columncolor{pink!30}}c}
\toprule[1.0pt]
Method~$\rightarrow$   & AE   & Ganomaly & f-AnoGAN & SALAD  & ProxyAno & SSD     & STPFM   & MKD     & PaDiM   & RD4AD   & AE-flow &  \\
Venue~$\rightarrow$     & Arxiv'18    & ACCV'18  & MIA'21   & TMI'21 & TMI'21   & ICLR'21 & BMVC'21 & CVPR'21 & ICPR'21 & CVPR'22 & ICLR'23 & \textbf{URA-Net}                         \\
Metric~$\downarrow$    & \cite{AE-SSIM}      & \cite{Ganomaly}          & \cite{f-anogan}          & \cite{SALAD}       & \cite{ProxyAno}         & \cite{sehwag2021ssd}        & \cite{STPM}         & \cite{MKD}        & \cite{PaDiM}        &  \cite{RD4AD}       &    \cite{AE-flow}       &                        \\ \midrule
AUROC    & 77.8 & 83.5     & 83.4     & 96.4   & 93.3     & 92.3    & 96.9    & 96.7    & 96.9    & 97.6    & \uline{98.1}    & \textbf{98.6}                     \\
F1-score & 85.8 & 88.7     & 84.7     & 93.4   & 72.5     & 91.9    & 95.8    & 94.6    & 95.2    & \uline{96.4}    & \uline{96.4}    & \textbf{97.1}                     \\
ACC      & 78.3 & 81.6     & 77.5     & 90.6   & 84.9     & 87.2    & 93.7    & 91.6    & 92.8    & \uline{94.6}    & 94.4    & \textbf{95.7} \\   
\bottomrule[1.0pt]
\end{tabular}}}
\end{table*}

\begin{figure}
    \centering
    \includegraphics[width=87mm]{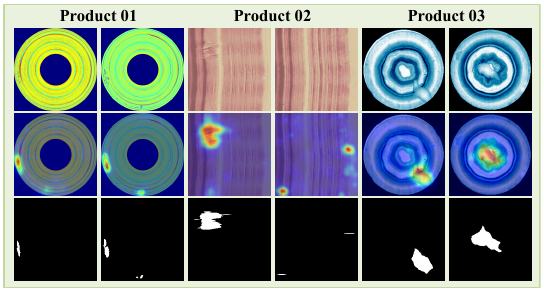}
    \caption{The localization results of the proposed URA-Net on the BTAD benchmark \cite{BTAD}.}
    \label{fig:BTAD}
\end{figure}

\subsection{Comparative Experiments}
\subsubsection{Experimental Results on MVTec AD}
To validate the superiority of our proposed method, we compared it with state-of-the-art (SOTA) methods published in the past three years, including embedding-based approaches such as PaDiM \cite{PaDiM}, PFM \cite{MBPFM}, PatchCore \cite{PacthCore}, CDO \cite{caoyunkangtii}, and ADPS \cite{ADPS}; image reconstruction methods like RIAD \cite{RIAD}, DRAEM \cite{draem}, and OCR-GAN \cite{draem}; and feature reconstruction methods such as UTRAD \cite{chen2022utrad}, RD4AD \cite{RD4AD}, and FOD \cite{fod}.\\
\indent The quantitative comparison results on MVTec AD are presented in Table \ref{table:mvtec}. The proposed method achieves the best overall average detection and localization performance, with an image-wise AUROC of 99.4\% and a pixel-wise AUROC of 98.5\%. Compared to the current SOTA feature reconstruction method, FOD, our method shows an improvement of +0.7\% in image-wise AUROC and +0.2\% in pixel-wise AUROC. This improvement stems from the ability of our proposed RAM to enable the model to utilize global normal semantic information for restoring anomalous features, whereas FOD lacks this explicit anomaly restoration mechanism. Additionally, it is noteworthy that our method achieves 100\% image-wise AUROC on the \textit{Grid}, \textit{Leather}, \textit{Tile}, \textit{Bottle}, \textit{Cable}, \textit{Hazelnut}, and \textit{Transistor} categories, further demonstrating its superior detection performance.\\
\indent The qualitative comparison results on the MVTec AD dataset are depicted in Fig. \ref{fig:mvtec}. Compared to other outstanding methods, our proposed approach achieves more precise localization of anomaly regions and effectively mitigates noise in background areas. Furthermore, our method exhibits outstanding detection performance across different sample categories, showcasing its robust generalization capability.

\subsubsection{Experimental Results on BTAD}
Anomaly detection and localization on BTAD pose significant challenges due to its complex texture background. Therefore, to further substantiate the effectiveness of our method, we conduct performance comparison experiments on the BTAD dataset.\\
\indent The quantitative experimental results are depicted in Table \ref{table:btad}. Our method attains the highest overall detection and localization performance, achieving an image AUROC of 96.0\% and a pixel AUROC of 97.6\%. Regarding image-level and pixel-level AUROC, URA-Net exceeds the current SOTA method FOD by +0.6\% and +0.1\%, respectively. 
Additionally, in the Product 01 and Product 03 categories, our method achieved close to 100\% image-level AUROC, further substantiating the superiority of our method.\\
\indent The qualitative experimental results are illustrated in Fig. \ref{fig:BTAD}. Even in complex textured backgrounds, our method accurately localizes anomaly regions across three product types, demonstrating the effectiveness of our approach.

\subsubsection{Experimental Results on OCT-2017}
To further validate the generalization capability of our method, we compare its performance with SOTA medical image anomaly detection methods, including AE \cite{AE-SSIM}, Ganomaly \cite{Ganomaly}, f-AnoGAN \cite{f-anogan}, SALAD \cite{SALAD}, ProxyAno \cite{ProxyAno}, SSD \cite{sehwag2021ssd}, STPFM \cite{STPM}, MKD \cite{MKD}, PaDiM \cite{PaDiM}, RD4AD \cite{RD4AD}, and AE-flow \cite{AE-flow}, on the OCT-2017 medical image dataset.\\
\indent The quantitative experimental results are presented in Table \ref{table:oct-2017}. Our method achieves the best detection performance, with an image AUROC of 98.6\%, an F1-score of 97.1\%, and an ACC of 95.7\%. Compared to the second-best results, our method improves by +0.5\% in image AUROC, +0.7\% in F1-score, and +1.1\% in ACC. This demonstrates that our method is applicable not only to industrial images but also to medical images, showcasing its robust generalization capability.\\
\indent Fig. \ref{fig:oct-2017} illustrates the qualitative detection results. Our proposed method effectively detects anomalies in medical images.

\begin{table*}[!ht]
\centering
\caption{\textbf{Ablation analysis} on the \textbf{MVTec AD} dataset \cite{MVTEC}. \textbf{IASM} refers to image-level artificial anomaly synthesis module. `\textbf{w/o SK}' indicates that removing skip-connection.} 
\label{table:ablation}
\vspace{-7pt}
\fontsize{10}{14}\selectfont{
\resizebox{0.9\textwidth}{!}{
\begin{tabular}{c|ccccc|ll|ccc}
\toprule[1.0pt]
Variant & IASM & FASM & UIAPM & RAM & w/o SK & I-AUROC & P-AUROC & FPS & Parameters & FLOPs \\ \midrule
A       &\textcolor{gray}{\xmark}     &\textcolor{gray}{\xmark}      & \textcolor{gray}{\xmark}      &\textcolor{gray}{\xmark}     & \textcolor{gray}{\xmark}       & 97.6        & 97.3        & 63.25    & 97285184       & 30598758400      \\
B       &\cmark      &\textcolor{gray}{\xmark}      &\textcolor{gray}{\xmark}       &\textcolor{gray}{\xmark}     &\textcolor{gray}{\xmark}        & 98.2\footnotesize{\textcolor{purple}{+0.6$\uparrow$}}        & 97.8\footnotesize{\textcolor{purple}{+0.5$\uparrow$}}        & 63.25    & 97285184       & 30598758400      \\
C       &\textcolor{gray}{\xmark}      &\cmark      &\textcolor{gray}{\xmark}       &\textcolor{gray}{\xmark}    &\textcolor{gray}{\xmark}        & 98.4\footnotesize{\textcolor{purple}{+0.8$\uparrow$}}        & 97.9\footnotesize{\textcolor{purple}{+0.6$\uparrow$}}        & 63.25    & 97285184       & 30598758400      \\
D       &\textcolor{gray}{\xmark}      &\cmark      &\cmark       &\textcolor{gray}{\xmark}     &\textcolor{gray}{\xmark}        & 98.8\footnotesize{\textcolor{purple}{+1.2$\uparrow$}}        & 98.1\footnotesize{\textcolor{purple}{+0.8$\uparrow$}}        & 57.11    & 97286722       & 30599151616      \\
E       &\textcolor{gray}{\xmark}      &\cmark      &\cmark       &\cmark     &\textcolor{gray}{\xmark}        & 99.2\footnotesize{\textcolor{purple}{+1.6$\uparrow$}}         & 98.4\footnotesize{\textcolor{purple}{+1.1$\uparrow$}}        & 55.10     &97286746        &30599151616       \\
\rowcolor{pink!30} 
F &\textcolor{gray}{\xmark}      &\cmark      &\cmark       &\cmark     &\cmark        & 99.4\footnotesize{\textcolor{purple}{+1.8$\uparrow$}}        & 98.5\footnotesize{\textcolor{purple}{+1.2$\uparrow$}}        & 55.10     & 97286746       &30599151616    \\
\bottomrule[1.0pt]
\end{tabular}}}
\end{table*}
\begin{figure}[!t]
    \centering
\includegraphics[width=87mm]{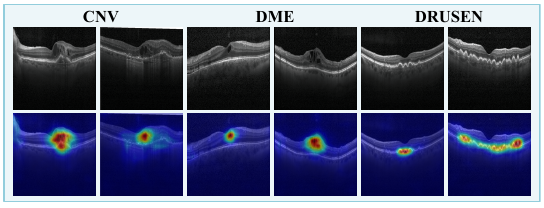}
    \caption{The localization results of the proposed URA-Net on the OCT-2017 medical dataset \cite{oct2017}.}
    \label{fig:oct-2017}
\end{figure}
\begin{figure}
    \centering
    \includegraphics[width=87mm]{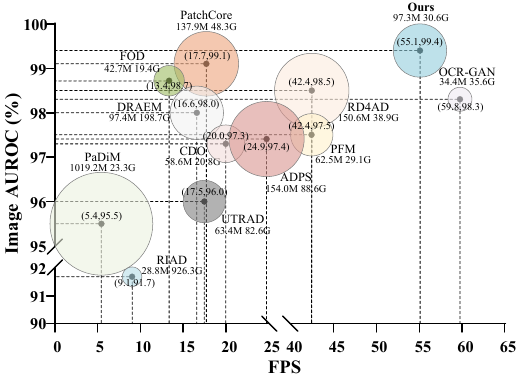}
    \caption{\textbf{Frames per second (FPS) versus image-wise AUROC} on MVTec AD benchmark \cite{MVTEC}. The size of the circles represents the number of parameters. The floating point operations (FLOPs) are also reported. ($\cdot$, $\cdot$) represents (FPS, Image AUROC).}
    \label{fig:real-time}
\end{figure}
\subsection{Complexity Analysis}
In real-world industrial settings, achieving a good balance between the detection accuracy and detection speed of models is crucial. Therefore, we conduct a complexity analysis of the proposed URA-Net and existing SOTA methods. Specifically, this study analyzes the complexity of the models from three perspectives: i) Frames Per Second (FPS), ii) the number of parameters, and iii) Floating Point Operations (FLOPs).\\
\indent As illustrated in Fig. \ref{fig:real-time}, our method attains the second fastest detection speed, registering an FPS of 55.1, only slightly behind OCR-GAN's 59.8. However, our method exhibits markedly superior detection accuracy compared to OCR-GAN. In contrast to the current SOTA feature reconstruction method FOD, our approach is approximately \textbf{4$\times$} faster. Compared to the current SOTA feature embedding method PatchCore (137.9M parameters, 48.3G FLOPs), our approach achieves superior detection performance with fewer parameters (97.3M) and FLOPs (30.6G). Additionally, our method achieves a detection speed approximately \textbf{3$\times$} faster than PatchCore. 
In summary, our approach achieves SOTA detection accuracy with fewer parameters and faster computational speed.

\begin{figure}
    \centering
    \includegraphics[width=87mm]{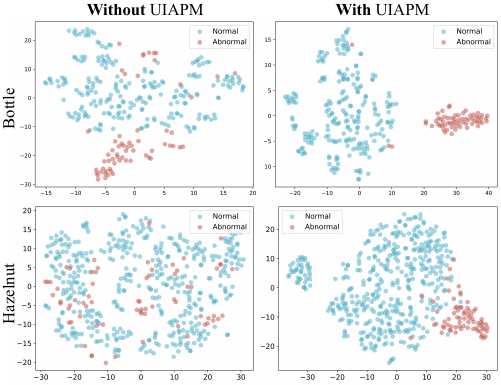}
    \caption{\textbf{Visualization of the impact of UIAPM}. We plot the t-SNE visualization of normal and abnormal instances for Bottle and Hazelnut categories in MVTec AD \cite{MVTEC} dataset.}
    \label{fig:TSNE}
\end{figure}
\subsection{Ablation Experiments}
In this section, we conduct further ablation experiments on MVTec AD to analyze the influence of the proposed modules. Initially, we utilize a baseline (Variant A in Table \ref{table:ablation}) for anomaly detection.
Then we progressively integrate the proposed modules into the baseline model, resulting in our comprehensive model (Variant F in Table \ref{table:ablation}), which achieves SOTA detection performance.
\subsubsection{Influence of FASM}
{FASM is proposed to assist the model in learning to restore anomalies by generating anomalies at the feature level. 
As shown in Table \ref{table:ablation}, compared to the model without FASM (Variant A), the model with FASM (Variant C) improves image-wise AUROC by +0.8\% and pixel-wise AUROC by +0.6\%, without any change in inference speed or parameter count. Additionally, existing methods typically construct artificial anomalies at the image level, which we refer to as the Image-level Artificial Anomaly Synthesis Module (IASM) in our study. To validate that our proposed FASM is superior to IASM, we replace FASM with IASM, creating model variant B. Without changing the parameter count or inference speed, the model with FASM (Variant C) improves image-wise AUROC by +0.2\% and pixel-wise AUROC by +0.1\% compared to the model with IASM (Variant B in Table \ref{table:ablation}). These experimental results not only demonstrate the effectiveness of our proposed FASM but also validate its superiority over IASM.}
\subsubsection{Influence of UIAPM}
We propose UIAPM to roughly estimate abnormal regions, laying the foundation for subsequent anomaly restoration in RAM. 
As indicated in Table \ref{table:ablation}, compared to the model without UIAPM (Variant C), the model with UIAPM (Variant D) shows improvements of +0.4\% in image-wise AUROC and +0.2\% in pixel-wise AUROC, with only a slight increase in the number of parameters and FLOPs, and a minor decrease in FPS. In addition, as illustrated in Fig. \ref{fig:TSNE}, without UIAPM, the model struggles to distinguish between normal and anomalous features. UIAPM employs discriminative learning to make the differences between normal and anomalous features more pronounced, thereby enhancing the model's ability to differentiate between them. These experimental results highlight the effectiveness and superiority of UIAPM.
\begin{figure}
    \centering
\includegraphics[width=\linewidth]{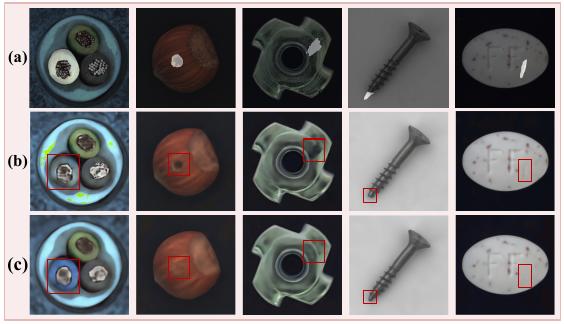}
    \caption{\textbf{Visualization of the impact of RAM}. (a) Input anomalous image and corresponding label. (b) Reconstruction result \textbf{without RAM}. (c) Reconstruction result \textbf{with RAM}. It is noteworthy that our proposed method relies on feature reconstruction. The reconstructed images are generated by training a decoder, which is exclusively employed
for visualization.}
    \label{fig:rec-result}
\end{figure}
\begin{figure}
    \centering
    \includegraphics[width=\linewidth]{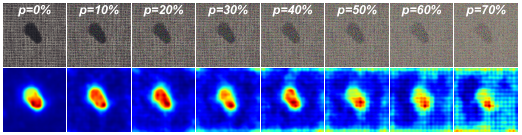}
    \caption{{\textbf{Localization results under different noise levels}. From top to bottom are the input noisy images and the corresponding detection results, respectively.}}
    \label{fig:noise}
\end{figure}
\subsubsection{Influence of RAM}
The objective of RAM is to leverage global normal semantic information to restore anomalous features, thereby boosting the overall performance of the model. 
As depicted in Table \ref{table:ablation}, compared to the model lacking RAM (Variant D), the model with RAM (Variant E) exhibits a +0.4\% improvement in image-wise AUROC and a +0.3\% enhancement in pixel-wise AUROC, highlighting the effectiveness of RAM. These improvements come with only a slight increase in the number of parameters and FLOPs, and a minor decrease in FPS. Fig.~\ref{fig:rec-result} provides a more intuitive illustration of the effectiveness of RAM. Without RAM, the model still reconstructs the anomalous regions; in contrast, with RAM, the model successfully repairs these anomalous regions.
\begin{figure}
    \centering
\includegraphics[width=\linewidth]{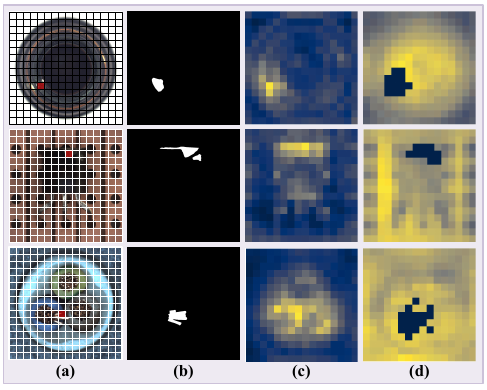}
    \caption{
\textbf{Comparison of attention map visualizations between the self-attention module (SAM) and the proposed RAM}. (a) Anomalous image, where each \textcolor{red}{red} rectangle represents an anomalous patch. (b) Corresponding label. (c) Attention map generated by the SAM. (d) Attention map generated by the proposed RAM.}
    \label{fig:atten_map}
\end{figure}
\begin{figure}[!t]
    \centering
\includegraphics[width=\linewidth]{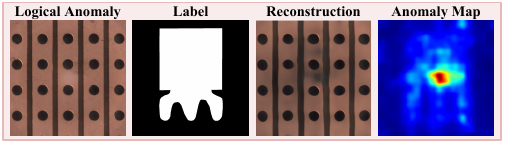}
    \caption{
\textbf{Analysis of failure case} of the proposed method.}
    \label{fig:failcase}
\end{figure}
\subsubsection{Influence of removing skip-connection}
In traditional Transformer blocks, skip connections are commonly employed. However, these skip connections can directly introduce anomalies into subsequent operations, leading to the persistence of anomalies in the reconstruction process. 
Therefore, in model variant F, we remove some skip connections present in model variant E. As shown in Table \ref{table:ablation}, in comparison to the model with skip connections (Variant E), the model without skip connections (Variant F) demonstrates enhancements of +0.2\% and +0.1\% in image-wise AUROC and pixel-wise AUROC, respectively, underscoring the significance of removing skip connections. Remarkably, these enhancements are accomplished without augmenting the model's parameter count, FLOPs, or diminishing FPS. 

\subsection{{Robustness against Noise}}
{In real-world industrial scenarios, it is essential for models to exhibit strong robustness against noise. To this end, we conduct noise robustness experiments on the proposed method. Specifically, as shown in the top row of Fig,~\ref{fig:noise}, we apply probabilistic speckle noise to corrupt the original images, where each pixel is replaced with a uniformly distributed random value with a probability $p$. The bottom row of Fig.~\ref{fig:noise} shows the detection results of our method under varying noise levels. Notably, even under a severe noise condition with $p=50\%$, our method can still accurately localize anomalies, highlighting its strong robustness to noise.}
\subsection{Attention Map Visualization}
To intuitively illustrate what our proposed URA-Net has learned, we visualize the attention maps. As shown in Fig. \ref{fig:atten_map}, in a vanilla self-attention module, anomalous features exhibit strong correlations with both themselves and adjacent anomalous features, leading to the reconstruction of these anomalies. In contrast, in our proposed RAM, anomalous features display strong correlations with global normal features and minimal association with themselves or neighboring anomalous features. This enables the effective restoration of anomalous features using global normal features.

\subsection{Analysis of Limitations}
Although our method achieves SOTA results on multiple datasets, it still has limitations in detecting logical anomalies. As shown in Fig.~\ref{fig:failcase}, in the \textit{Transistor} category, due to the ``misplaced" anomaly, the model fails to capture the semantic information of the transistor, resulting in an incomplete reconstruction and hence a missed detection. In future work, we will explore using a normal image as a prompt combined with RAM to enhance the model's ability to detect logical anomalies.

\section{Conclusion}

In this study, we introduce a novel unsupervised anomaly detection method, URA-Net, which explicitly guides the anomaly restoration process to enhance restoration quality. By constructing artificial anomalies at the feature level, FASM aids the model in learning to restore anomalies. Additionally, UIAPM is proposed to roughly estimate anomalous regions, laying the foundation for subsequent anomaly restoration. Furthermore, we introduce RAM, which leverages global normal semantic information under the guidance of UIAPM to restore detected anomalous features, thus enhancing the quality of anomaly restoration and overall model performance. Extensive experiments on three public datasets demonstrate the superior anomaly detection performance of our proposed method. In future research, we plan to extend URA-Net to multi-class anomaly detection.


\bibliographystyle{IEEEtran}
\bibliography{ref}

\end{document}